\documentclass[10pt,twocolumn,letterpaper]{article}
\UseRawInputEncoding
\usepackage{cvpr}
\usepackage{times}
\usepackage{epsfig}
\usepackage{graphicx}
\usepackage{multirow}
\usepackage{amsmath}
\usepackage{subfigure}
\usepackage{amssymb}
\usepackage{booktabs}
\usepackage{widetext}
\usepackage{multicol,lipsum}
\usepackage[breaklinks=true,bookmarks=false]{hyperref}

\cvprfinalcopy 

\begin{document}

\title{Tracklets Predicting Based Adaptive Graph Tracking}

\author{
Chaobing Shan$^{1,2}$, Chunbo Wei$^2$, Bing Deng$^2$, Jianqiang Huang$^2$, Xian-Sheng Hua$^2$\\
Xiaoliang Cheng$^{1}$, Kewei Liang$^{1,}\thanks{Corresponding Author}$\\
$^{1,\ast}$ School of Mathematical Sciences, Zhejiang University \ \ \ \ \ \ $^2$DAMO Academy, Alibaba Group\\
chaobing\_s@zju.edu.cn, chunbo.wcb@alibaba-inc.com, dengbing.db@alibaba-inc.com\\
jianqiang.hjq@alibaba-inc.com, huaxiansheng@gmail.com, xiaoliangcheng@zju.edu.cn, matlkw@zju.edu.cn}

\maketitle
\begin{abstract}
Most of the existing tracking methods link the detected boxes to the tracklets using a linear combination of feature cosine distances and box overlap. But the problem of inconsistent features of an object in two different frames still exists. In addition, when extracting features, only appearance information is utilized, neither the location relationship nor the information of the tracklets is considered. We present an accurate and end-to-end learning framework for multi-object tracking, namely \textbf{TPAGT}. It re-extracts the features of the tracklets in the current frame based on motion predicting, which is the key to solve the problem of features inconsistent. The adaptive graph neural network in TPAGT is adopted to fuse locations, appearance, and historical information, and plays an important role in distinguishing different objects. In the training phase, we propose the balanced MSE LOSS to successfully overcome the unbalanced samples. Experiments show that our method reaches state-of-the-art performance. It achieves 76.5\% MOTA on the MOT16 challenge and  76.2\% MOTA on the MOT17 challenge.
\end{abstract}

\section{Introduction}\label{section:introduction}

Object tracking is a very important task in computer vision.
It has many applications in autonomous driving, video surveillance, behavior prediction, traffic management, and accident prevention\cite{Keni2008,luo2014multiple}.
The multi-object tracking (MOT) always deals with various objects, such as pedestrians, vehicles, athletes, animals, and so on.
Each object is given a unique id (identity) and the trajectory is formed. A new object will be given a new id, while a disappearing objects need to be removed from the tracklets.
Compared with single-object tracking, MOT is more complicated, which is mainly used in video scenes.
Many videos cover different scenes, such as sharpness range from clear to blurry, light intensity range from the day to the night, viewing angle range from high to low, and camera lenses range from static to moving. In addition, many objects in the video are occluded, or often reflected in mirrors or windows, or will only be visible at the edge of the camera.
The appearance of different objects is often similar, and they may be deformed, too small in size or very densely located, etc.
What is more, when an object is in motion, its posture, video shooting angle and even the light intensity may change.
 All of these problems are more complicated for multi-object tracking\cite{Milan2016MOT}.

A common practice in \cite{Bewley2016,Wojke2017,Liu2018,Fu2019,Sergey2019,Hou2019,Maher2018} is using the Kalman filter\cite{Welch1995} to predict the locations of the tracklets in the current frame, then the Hungarian algorithm\cite{H1995} is adopted for matching. In some existing methods \cite{zhou2020tracking,Caelles2016,Z2017,Malinovskiy2011,Mittal2002,Segen1996,Cohen2018,Khan2009}, CNN or variational method is selected to estimate the optical flow of the tracklets from the previous frame to the current frame, and the greedy algorithm is adopted to complete the matching. In the literature \cite{zhang2020simple,zhang2018integrated,lu2020retinatrack}, a more refined feature vector is extracted by
an adding re-id branch, a similarity matrix is formed by a linear combination of feature similarity and IOU(Intersection over Union), and the matching is completed by the Hungarian algorithm.  Although these methods have made great progress in multi-object tracking, there are still the following
limitations.
\begin{enumerate}
    \item The feature of the tracklets is not from the current frame. During the movement, an object may change its posture.
    The light intensity and video shooting angle may change too.
    These lead to inconsistent features of 
    an object extracted from different frames.
    The tracking accuracy is severely reduced during matching.
    \item 
    In extracting feature, only appearance information is utilized. Neither the location relationship nor the information of the tracklets is  considered.
    \item 
    There exist a drawback of unbalanced samples.
    A tracklet can only match a detected bbox(bounding box) 
    that is a continuous positive sample. The rest unmatched tracklets
    belong to continuous negative samples.
    Obviously, the number of continuous positive samples is less than that of continuous negative samples.
    More worsely, few new objects and disappearing objects are observed.
    Therefore, the sample numbers of the 
    different types are unbalanced.
\end{enumerate}

To solve the above problems, we need to extract the features of the tracklets in the current frame. But, we cannot use the bboxes of the tracklets directly, because they are not aligned with the detected bboxes.
We predict the motion of the tracklets 
and re-extract their features
in the current frame.
Note that the connection between different
objects
look like a graph network. Each object is as a node, 
and the information can be spread through the connection between nodes.
This motivates us to adopt the graph neural networks (GNN)\cite{Scarselli2009} to integrate more information of tracklets and obtain better features of objects.
In order to further maintain the global spatiotemporal location, appearance information, and historical information of each object, our GNN, termed adaptive graph neural networks (AGNN), needs to learn itself adaptive weights.

Consequently, in this paper, we 
present an end-to-end learning framework based on GNN, termed TPAGT tracker.
TPAGT
uses the sparse optical flow method to calculate the center location of tracklets in the current frame. 
ROI Align\cite{He2018ROIAlign} and fully connected layers are adopted to extract the initial appearance feature vectors of the tracklets and the detected objects. 
We input the feature vector into the graph neural network and get the similarity matrix of the detected objects.
Moreover, in the training phase, we propose the balanced MSE LOSS to successfully overcome the unbalanced samples. Experiments show that our method reaches state-of-the-art performance. It achieves 76.5\% MOTA on the MOT16 challenge and  76.2\% MOTA on the MOT17 challenge.

The main contributions of our method are:
\begin{enumerate}
  \item Trackle prediting based feature re-extraction. We predict the motion of tracklets, and re-extract their features in the current frame so that the features of the tracklets can be aligned with the features of the detected objects in the current frame. This causes in high probability to get the consistent features of an object from different frames.
  \item AGNN.
  Through integrating the locations and appearance information of tracklets and the detections, AGNN can update features and re-identify occluded objects.
  \item BMSE.
  The balanced MSE Loss is proposed to overcome the challenge of unbalanced  samples.
  \item SOTA. TPAGT significantly outperforms existing state-of-the-art methods
  both on public and private detections of MOT Challenges.
\end{enumerate}

The following contents in this article are organized as follows: Section 2 reviews the related work about the tracking. In Section 3, we introduce the TPAGT tracker. Section 4 shows the main results and ablation experiments. The summary is listed in Section 5.

\section{Related Work}\label{section:Related Work}
This paper divides existing work into two categories according to whether feature matching is adopted.

\textbf{Featureless matching: }SORT\cite{Bewley2016} detects the objects' locations and categories by Faster-RCNN\cite{He2015}, uses the Kalman filter\cite{Welch1995} to predict the tracklets' location in the current frame, and calculates the IOU, finally applies the Hungarian algorithm to link detected boxes to the tracklets. CenterTrack\cite{zhou2020tracking} inherits the network structure of CenterNet\cite{zhou2019objects}.
It has four more channels on the input than CenterNet: the 3-channel RGB image of the previous frame and the heatmap after the maximum pooling on the category channel. The output has one more branch: the optical flow prediction branch, which is used to predict the objects' displacement from the previous frame to the current frame. Finally, the distance between the tracklets and the detected objects is calculated, and a simple greedy algorithm is used to complete the matching.

\textbf{Feature matching:}
featureless matching actually looks at the location relationship between the tracklets and the detected objects. In addition to the important factor of location, the appearance of the object is also a very important factor, such as black clothes, white clothes, fat, thin, front, back, and so on are all important feature that distinguish different objects. RetinaTrack\cite{lu2020retinatrack} corrects the shortcoming that the structure of RetinaNet\cite{Linfocal2017} is not suitable for capturing the features of each instance. The more specific instance features corresponding to the anchor shape are captured before the classification and border regression, and then add a feature embedding branch, which is used for calculating feature distance, then track. FairMOT\cite{zhang2020simple} proposed the reasons why the features captured by the anchor-based model are not good.
However, the Re-id output of FairMOT is a one-hot vector, When the actual number of objects is much more than the output dimension, it will not be able to track.
For the first time, Wang\cite{wang2020joint} et al. proposed a framework for joint detection and tracking using GNN. The detected objects and tracklets are divided into bipartite graphs.
There is a problem here, in the graph neural network, the weights of the edges between the detected objects and the tracklets are the same. This may result in the failure to learn good features, or the final feature vectors are averaged, that is, all the feature vectors of the objects are almost the same.

In addition to tracking based on detection, there is also tracking to facilitate detection. Zhang\cite{zhang2018integrated} et al. proposed the Tracklet-Conditioned formula. Based on the detection results of Faster-Rcnn and the known historical trajectories, the bayesian formula is fully utilized to update the probability of the detection category. One more feature embedding branch is added to Faster-Rcnn architecture to calculate the cosine similarity between objects' feature vectors, and then matching. But this method also has drawbacks, because the Tracklet-Conditioned formula may only improve the accuracy of the classification, but not the accuracy of the bounding box coordinates.

In order to obtain a good tracking result, it makes sense that the location, appearance, and tracklets information is fully utilized.  Although some of the above work predicts the location of the tracklets in the current frame, it does not use the predicted location to update the tracklets' feature vectors. And when an object is occluded, a small part of the occluded object will obviously not be occluded, we need give that small part a large weight to get a more refined Re-id feature vector. Besides, an object reappears after being occluded, it is necessary to aggregate the global spatiotemporal information to update the Re-id feature vector. In addition, all these work ignored the unbalanced distribution of four types of samples.

\section{The TPAGT Architecture}\label{section:TPAGT Architecture}
\begin{figure*}[!htp]
	\centering
	\includegraphics[width=1.0\linewidth]{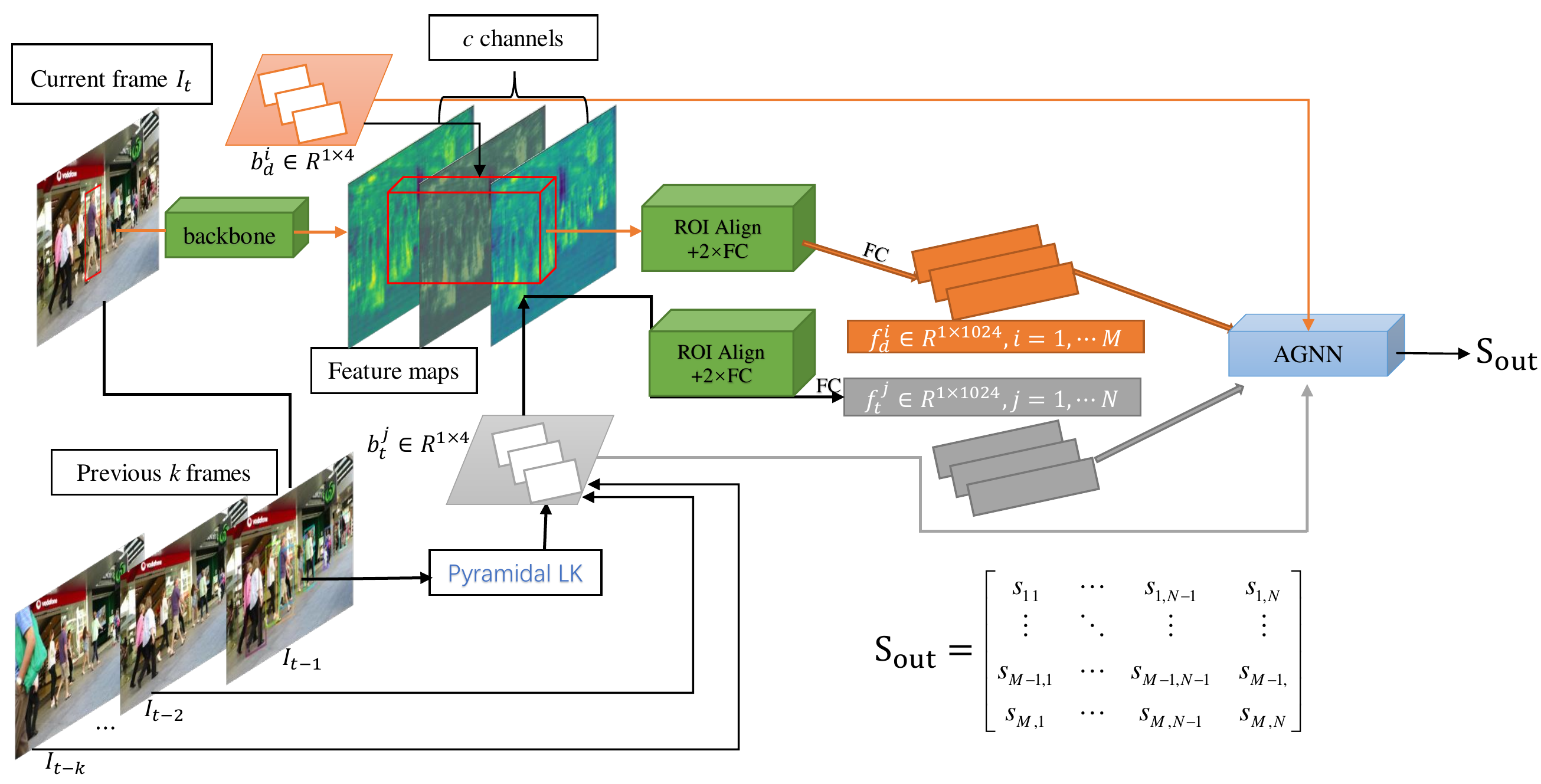}
	\caption{TPAGT Frame-work.}
	\label{fig:framework}
\end{figure*}

As shown in Figure \ref{fig:framework}, the TPAGT is an end-to-end tracking method, which consists of two main steps. The first step is tracklets predicting based feature re-extracting. It extracts features of detected objects and re-extracts features of tracklets in the current frame. The second step is updating features in adaptive graph tracking. The adaptive graph neural network is adopted with integrating all temporal and spatial information of the tracklets. Obviously, this is better than extracting only with the appearance information. We also introduce the balanced MSE Loss in training time for balancing sample distribution. Finally, we adopt the Hungarian algorithm in an augmented similarity matrix to complete matching in inference time.


\subsection{Tracklets predicting based feature re-extracting}
\begin{figure*}[htbp]
  \subfigure[bbox from $I_{t-1}$]{
   \begin{minipage}[t]{0.32\linewidth}
   \centering
   \includegraphics[width=4cm,height=3cm]{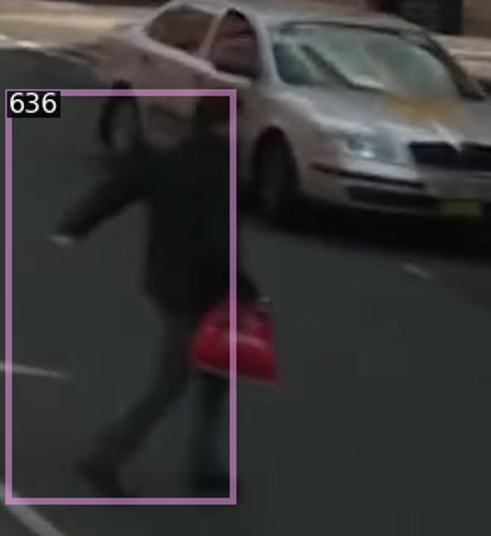}
   \end{minipage}}%
   \hfill
  \subfigure[predicted bbox]{
   \begin{minipage}[t]{0.32\linewidth}
   \centering
   \includegraphics[width=4cm,height=3cm]{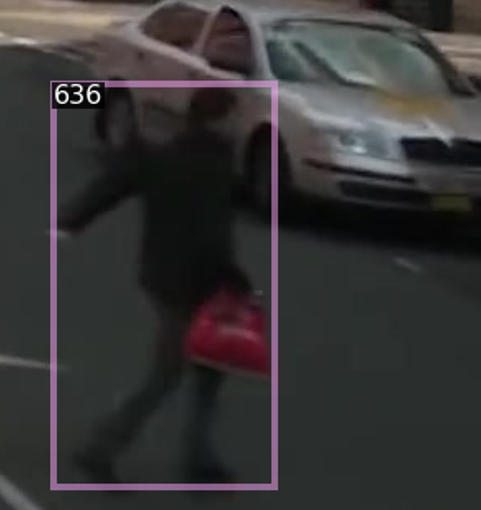}
   \end{minipage}}%
   \hfill
  \subfigure[ground truth bbox]{
   \begin{minipage}[t]{0.32\linewidth}
   \centering
   \includegraphics[width=4cm,height=3cm]{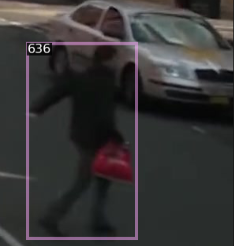}
   \end{minipage}}
   \caption{Benefits of using PyraLK. All three pictures contain the same object in the current frame $I_t$. The bbox in (a) is from the previous frame $I_{t-1}$. (b) uses the pyramid LK algorithm to predict the motion of the object, so the object's center position in the current frame can be found(bbox's width and height remain unchanged), and (c) is the ground truth bbox. We can se that (b) and (c) is almost aligned.}
   \label{fig:PyrLK}
\end{figure*}

As mentioned in Section \ref{section:introduction}, we use the sparse optical flow method, cause we only need to predict the motion of bboxes' center points. In some cases, the speed of an object is very large, we must choose the suitable method of optical flow for large motion estimation, so we apply the pyramid LK algorithm \cite{Bouguet00pyramidalimplementation}. As shown in the figure\ref{fig:PyrLK}, even if the object has a large motion between the previous and current frames, it can still align the tracklet bbox with the detected bbox well when using the pyramid LK algorithm.

The current frame $I_t$ is transformed into feature maps by backone, the detected $M$ bboxes use ROI Align to extract region features. These features are transformed into feature vectors through fully connected layers. At the same time, the pyramid-LK algorithm predicts the locations of tracklets(from the $I_{t-1}$ frame) in the current frame, with the bboxes of the previous $k$ frames (except for the $I_{t-1}$ frame), a total of $N$ historical bboxes are also transformed into feature vectors. The bboxes and feature vectors are the input of the adaptive graph neural network introduced in the next subsection. Here we re-extract the features of tracklets in the current frame, this can solve the problem of dissimilar features due to the large movement of the same object between different frames

\subsection{Adapted Graph Neural Network}
We treat detected objects and tracklets as a bipartite graph. Each object or tracklet is as a node. Each detected object has a connection with all tracklets, but there is no connection between any two detected objects, nor two tracklets. The learning goal of GNN is obtaining the perception information of each node $v$'s (perception is relative to compression, using existing information features and weights to reconstruct more complete information features) hiding state $\mathbf{h}_v(state\ embedding)$. For each node, its hidden state contains information of the neighboring nodes. We use the update formula:
\begin{equation}
\begin{aligned}
  h_{d,c+1}^{i}&=f\big(h_{d,c}^{i},\{ h_{t,c}^{j},e_{d,c}^{i,j} \}_{j=1}^N\big), \ i=1,2,\cdots,M,\\
  h_{t,c+1}^{j}&=f\big(h_{t,c}^{j},\{ h_{d,c}^{i},e_{t,c}^{j,i} \}_{i=1}^M\big)  \ j=1,2,\cdots,N.
\end{aligned}
\end{equation}
Here $f$ is the neural network, $h_{d,c}^{i}$ is the hidden feature vector of the i-th  detected object in the c-th layer, and $h_{t,c}^{j}$ is the hidden feature vector of  the j-th tracklet in the c-th layer. When $c=0$, $h_{d,0}^{i}=f_{d}^{i}, h_{t,0}^{i}=f_{t}^{j}$. $e_{d,c}^{i,j}$ represents the weight of the edge between the i-th detected object and the j-th tracklet in the $c$ layer. In this paper, we use the GNN of one hidden layer and add an adaptive part that will be introduced in the following.

When using AGNN(Adaptive Graph Neural Network) to update the feature vectors of all objects, we use the existing location and feature prior information as the weight of the edge $\mathrm{E}=[e^{i,j}]$ to aggregate feature vectors then updating them, instead of simply setting some unknown parameters to update the features. The specific aggregation steps are as follows:
\begin{enumerate}
  \item [1)] Calculate the initial feature similarity matrix
    \begin{equation}
        s_{i, j}=\frac{1}{\left\|f_{d}^{i}-f_{t}^{j}\right\|_{2}+1 \times 10^{-16}},
    \end{equation}
    \begin{equation}
        s_{i, j}=\frac{s_{i, j}}{\sqrt{s_{i, 1}^{2}+s_{i, 2}^{2}+\cdots s_{i, j}^{2}+\cdots+s_{i, N}^{2}}},
    \end{equation}
    \begin{equation}
        \mathbf{S}_{\mathrm{ft}}=\left[s_{i, j}\right]_{M \times N}, \ i= 1, \cdots M, j=1, \cdots N.
    \end{equation}
  \item [2)] Calculate the bbox IOU and form a prior similarity matrix from the result of step 1,
      \begin{equation}
        \mathrm{E}=w\times \mathrm{IOU}+(1-w)\times \mathbf{S}_{\mathrm{ft}}.
      \end{equation}
    The parameter $w$ measures the relative importance between locations information and appearance information, the neural network is good at learning it, the initial value is set to 0.5.
  \item [3)] Aggregation features with adaptive weights,
      \begin{equation}\mathbf{F}_{\mathrm{t}}^{\mathrm{ag}}=\mathrm{E} \mathbf{F}_{t}=\mathrm{E}\left[f_{t}^{1}, f_{t}^{2}, \cdots, f_{t}^{N}\right]^{T},\end{equation}
    Then update the features,
     \begin{equation}
          \mathbf{H}_{\mathrm{d}}=\sigma\left(\mathbf{F}_{d} W_{1}+\operatorname{Sigmoid}\left(\mathbf{F}_{d} W_{a}\right) \odot \mathbf{F}_{\mathrm{t}}^{\mathrm{ag}} W_{2}\right),
    \end{equation}
    \begin{equation}
      \mathbf{H}_{\mathrm{t}}=\sigma\left(\mathbf{F}_{t} W_{1}+\operatorname{Sigmoid}\left(\mathbf{F}_{t} W_{a}\right) \odot \mathbf{F}_{\mathrm{d}}^{\mathrm{ag}} W_{2}\right).
      \end{equation}
  Where $\odot$ is the dot product. The values in different dimensions of an object's feature vector represent the important information captured from this object's different parts. And different parts may be the key to distinguishing the objects. Therefore, when aggregating features, We need to multiply the values of different dimensions by different weights, the weights need to be determined by the input feature vector, so we call $W_a$ as the adaptive parameter, and $\operatorname{Sigmoid}\left(\mathbf{F}_{d} W_{a}\right)$ as the adaptive weight. That's the reason why we call it the adaptive graph neural network.
\end{enumerate}

The existing graph network tracking algorithm needs additional fully connected layers to reduce the dimension of vectors outputting from graph neural network, then calculates the euclidean distance to measure the similarity between the feature vectors. We only need normalize the feature vectors outputting from a simple one hidden layer graph neural network, and adopt the simple matrix multiplication to get the similarity matrix:
\begin{equation}
      h_d^i=\frac{h_d^i}{\| h_d^i \|_2},\ h_t^j=\frac{h_t^j}{\| h_t^j \|_2}, \ \mathbf{S}_{\text{out}=\mathbf{H}_{\mathrm{d}}\mathbf{H}_{\mathrm{t}}^{\mathrm{T}}}
\end{equation}
The output values range from 0 to 1. The larger the value, the more similar the two objects are. The purpose is to make the feature vectors of the same target close to coincide, and the feature vectors of different targets close to vertical. Its essence is equivalent to Triplet Loss\cite{lu2020retinatrack}, but simpler than Triplet Loss.
\subsection{Blanced MSE Loss}
\begin{figure}[!htp]
	\centering
	\includegraphics[width=1.0\linewidth]{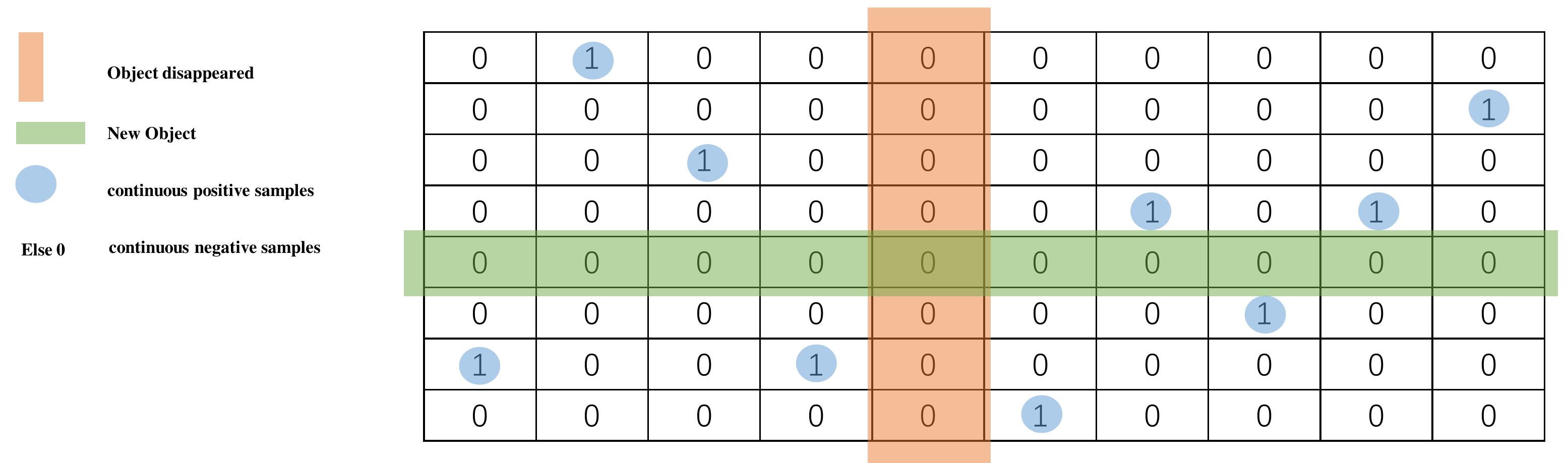}
   \caption{Explanation of unbalanced sample number distribution.}
	\label{fig:bmse}
\end{figure}

After outputting the similarity matrix,
we expect the lable corresponding to the same target to be 1,
and different targets to be 0.
As seen in figure \ref{fig:bmse}, each row presents a detected object in the current frame,
each column presents a tracklet in the previous frame.
The elements in the red rectangle column are all zero,
which means the historical object does not match any detected object,
so it is a disappeared object.
Elements in the green rectangle column are all zero,
which means the detected object does not match any historical object,
so it is a new appeared object which will be added a new id.
Apart from the above samples, the element which equals 1 means
they are the same object, so it is a pair of continuous positive sample.
An element that equals 0 means they are different objects,
so it is a pair of continuous negative sample.

When calculating loss, we use MSE Loss. However, most targets appear continuously, with only a small number of new and disappearing targets. In addition, each tracklet has at most one positive example in the current frame (label=1), the others are all negative examples (label=0), so the number of samples is extremely unbalanced. Therefore, we multiply different coefficients before the loss function corresponding to the new targets, disappearing targets, the targets of continuous positive and negative samples to balance the number distribution, and record the Loss as \textbf{blanced MSE Loss}(see equation \ref{EQ:BMSELOSS}).
\begin{widetext}
   \begin{equation}\label{EQ:BMSELOSS}
      \begin{aligned}
      \mathcal{L}=&\alpha E_{c 0}+\beta E_{c 1}+\gamma E_{ne}+\delta E_{d}+\varepsilon E_{w} \\
      =&\sum_{i=1}^{M} \sum_{j=1}^{N}\left[\begin{array}{c}
      \alpha\left(\hat{S}_{i, j}-S_{i, j}\right)^{2} \cdot \mathbb{I}_{\text {continue}} \cdot \mathbb{I}_{S_{i, j}=0}+\beta\left(\hat{S}_{i, j}-S_{i, j}\right)^{2} \cdot \mathbb{I}_{\text {continue}} \cdot \mathbb{I}_{S_{i, j}=1} \\
      +\gamma\left(\hat{S}_{i, j}-S_{i, j}\right)^{2}
      \cdot \mathbb{I}_{n e w}+\delta\left(\hat{S}_{i, j}-S_{i, j}\right)^{2} \cdot \mathbb{I}_{\text {disap}}+\varepsilon\|W\|_{2}^{2}
      \end{array}\right]
      \end{aligned}
   \end{equation}
\end{widetext}
Where $\mathbb{I}_{c}\left(S_{i, j}\right)=\left\{\begin{array}{c}1, \text { if } S_{i, j} \text { is the c target } \\ 0, \text { if } S_{i, j} \text { isn't the c target }\end{array}\right.$, $\alpha,\beta,\gamma,\delta,\varepsilon$ are the hyperparameters.

\subsection{Inference}
In the inference time, after the similarity matrix $\mathbf{S}_{\text {out}}$ is obtained through the adaptive graph neural network, we add a matrix of size $M\times M$, where all elements are equal to $margin =\pi$, the matrix is placed to the right side of $\mathbf{S}_{\text {out}}$ to form an augmented matrix, i.e., $\mathbf{S}_{\text {out}}=[\mathbf{S }_{\text {out}},\pi\times\mathbf{1}_{M\times M}]$, and then use the Hungarian algorithm to get the best match.

\begin{figure}[!htp]
	\centering
	\includegraphics[width=1.0\linewidth]{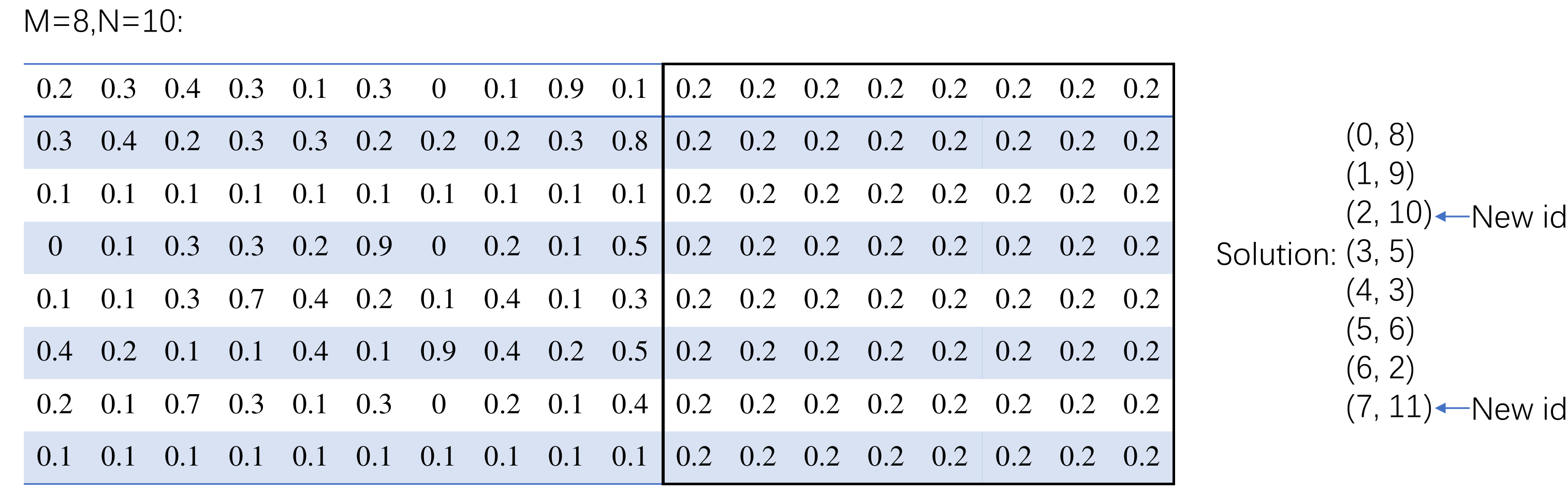}
	\caption{An example of our method. After Hungarian algorithm, we can get the 3rd and 8th objects are new objects, so their id are $\max\{id\}+1$ and $\max\{id\}+2$, respectively.}
	\label{fig:example}
\end{figure}

\textbf{Association match and new object appearing. }Assuming that one output pair of Hungarian Algorithm is $(i,j)$, $i$ is the number of rows, and $j$ is the number of columns. If $j<N$, then $i$ and trackle $j$ will be matched, and the id of $j$ is assigned to the object $i$. othreise, $i$ is a new object, then the id of the object $i$ is equal to $\max\{id\}+1$.
As shown in figure \ref{fig:example}, take $M=8,N=10$, and take margin $\pi=0.2$. According to the Hungarian algorithm, we can get that the third and the eighth objects in the solution are new objects, other objects are matched, the 3rd and 8th objects add 1 and 2 to the current max id number, respectively.

\textbf{Object disappearing. }We take $k=10$ in inference time. It means that the information of the object in the previous 10 frames is retained and is matched with the objects in the current frame. If an object in one frame does not match all objects in the next ten frames, the object is considered to disappear.

\section{Experiments}\label{section:Experiments}
\subsection{Datasets and Metrics}
\textbf{Datasets}
We evaluate our TPAGT on the testing sets of MOT16 and MOT17 benchmark. 
A brief review of MOT public datasets lists in Table \ref{tab:MOTPubData}. MOT17 includes 7 training videos and 7 testing videos, MOT16 contains same videos with MOT17. The MOT16 datasets provide public detections of the DPM detector, while the MOT17 provides public detections of three detectors (DPM, Faster-RCNN, and SDP). We perform the tracking over MOT testing sets and main results are reported on that. For private detection, we choose the detection results of the Fair\cite{zhang2020simple} for tracking. For ablation experiments, cause the MOT training sets provide the ground truth and the testing sets do not provide the ground truth, we divide the MOT17 training sets into two parts, half of which was used for training, and the other half as the validation set.
\begin{table}
\scriptsize
	\centering
   \renewcommand\tabcolsep{3.5pt}
   \begin{tabular}{c|c|cccccc}
   \toprule  
   \multicolumn{2}{c}{Dataset} & Videos & Detector & Identities & Boxes & Frames\\
   \midrule  
   MOT16& train & 7 & DPM & 517 & 110407 & 5316\\
   & test & 7 & DPM & 756 & 182326 & 5919 \\
   \midrule
  MOT17& train&7& DPM,FRCNN,SDP & 1638 & 336891 & 15948\\
   & test & 7 & DPM,FRCNN,SDP & 2355 & 564228 & 17757\\
   \bottomrule 
   \end{tabular}
   \caption{MOT datasets with public detection.}
   \label{tab:MOTPubData}
\end{table}

\textbf{Metrics }
We use the $7$ most widely used 
metrics 
to evaluate the quality of our tracker.
The main metric is the multi-object tracking accuracy (MOTA) \cite{Bernardinclear2008}
\begin{equation}
    MOTA=1-\frac{\sum_t(FP_t+FN_t+IDSW_t)}{\sum_tGT_t},
\end{equation}
where the subscript $t$ is the frame index and $GT$ 
is the number of ground truth bbox.
The second is the IDF1 Score\cite{Ristani2016}, 
the ratio of correctly identified detections
over the average number of ground truth and computed detections.
The third is the mostly tracked targets (MT), 
the ratio of ground-truth trajectories 
covered by a track hypothesis for at least 80\% of their respective life span.
The fourth one is the mostly lost targets (ML), 
the ratio of ground-truth trajectories 
covered by a track hypothesis for at most 20\% of their respective life span.
The remain three metrics are the number of false positives (FP), the number of false negatives (FN) and the number of identity switches (IDSW).

\subsection{Implementation Details}
{
We use ResNet101-FPN as backbone to perform main results. Four different architectures of neural networks, VGG16, ResNet34, ResNet50, and ResNet101(with or without FPN) are used as the backbone in the ablation experiment.
Note that these four neural networks are pre-trained on the coco data set.
After $30$ epochs of fine-tunings on the detecting task of the MOT datasets, their neural network parameters are no longer updated.
}

In data preprocessing, We adjust the sizes of the videos in MOT. Six videos are reduced from $1920\times 1080$ to $1333\times 750$, and the other $640\times 480$ video is resized to $800\times 450$.


In the pyramid LK algorithm, the neighborhood window we choose is $120\times120$, the number of iterations is 10, and the convergence criterion is $0.01$. The dimension of the output feature vector in the AGNN is $256$.
The hyper-parameters in balanced MSE loss
$\alpha=25,$ $\beta=1,$ $\gamma=50,$ $\delta=50,$ $\varepsilon=0.01$.
The learning rate $lr$ of neural networks is updated by the cosine annealing algorithm in the Adam optimizer from the initial $0.05$ to the final $2.5\times10^{-7}$.
It is updated every $30$ epochs.
The total epochs are $3000$. Our module is trained on 2 GPUS of Tesla-V100, with 32 GB memory in each GPU.


\subsection{Main Results}
\textbf{Public Detection Result }
The backbone we choose is ResNet101-FPN. From the table \ref{tab:MOTPub}, we can get that whether it is on MOT16 or MOT17 testing sets, our MOTA, MT, ML, FN metrics are almost best among all the results, where MOTA metric achieves 62.7\%, exceeds Unsup by 0.3 points on the MOT16; It achieves 62.0\%, exceeds Unsup by 0.3 points and CenterTrack by 0.6 points on the MOT17. The IDF1 metric is also almost optimal, it achieves 60.3\%, exceeds Unsup by 1.8 points on the MOT16; It achieves 62.0\%, exceeds Unsup by 1.4 points and CenterTrack by 5.2 points on the MOT17.
\begin{table}
\scriptsize
	\centering
   \renewcommand\tabcolsep{3.5pt}
   \begin{tabular}{c|ccccccc}
   \toprule  
   \multicolumn{8}{c}{MOT16}\\
   \midrule
   Method& MOTA$\uparrow$ & IDF1$\uparrow$ & MT$\uparrow$  & ML$\downarrow$ & FP$\downarrow$ & FN$\downarrow$ & IDSw$\downarrow$\\
   \midrule  
   RAR16\cite{FangPAR2018}& 45.9& 48.8 & 13.2 & 41.9 & 6871 & 91173 & 648\\
   AMIR\cite{SadeAMIR2017}& 47.2& 46.3 & 14.0 & 41.6 & \textbf{2681} & 92856 & 774\\
   MOTDT\cite{ChenMOT2018}& 47.6& 50.9 & 15.2 & 38.3 & 9253 & 85431 &792\\
   STRN\cite{XuSTRN2019}& 48.5& 53.9 & 17.0 & 34.9 & 9038 & 84178 &747\\
   KCF\cite{ChuKCF2019}& 48.8& 47.2 & 15.8 & 38.1 & 5875 & 86567 &906\\
   Tracktor\cite{Philipp2019}& 54.5& 52.5 & 19.0 & 36.9 & 3280 & 79149 & 682\\
   DeepMOT\cite{Xudeep2020}& 54.8& 53.4 & 19.1 & 37.0 & 2955 & 78765 &645\\
   MPNTrack\cite{GuMPN2020}& 58.6& 61.7 & 27.3 & 34.0 & 4949 & 70252 &\textbf{354}\\
   Lif\_T\cite{Hornakova2020}& 61.3& \textbf{64.7} & 27.0 & 34.0 & 4844 & 65401 &389\\
   Unsup\cite{KarUnsup2020}& 62.4& 58.5 & 27.0 & 31.9 & 5909 & 61981 &588\\
   ours& \textbf{62.7} & 60.3 & \textbf{28.5} & \textbf{26.9} & 5077 & \textbf{61952} &978\\
   \toprule  
   \multicolumn{8}{c}{MOT17}\\
   \midrule
   Method& MOTA$\uparrow$ & IDF1$\uparrow$ & MT$\uparrow$  & ML$\downarrow$ & FP$\downarrow$ & FN$\downarrow$ & IDSw$\downarrow$\\
   \midrule
   DMAN\cite{ZhuDMAN2018}& 48.2& 55.7 & 19.3 & 38.3 & 26218 & 263608 &2194\\
   STRN\cite{XuSTRN2019}& 50.9 & 56.0 & 18.9 & 33.8 & 25295 & 249365 &2397\\
   MOTDT\cite{ChenMOTDT2018}& 50.9& 52.7 & 17.5 & 35.7 & 24069 & 250768 &2474\\
   Tracktor17\cite{Bergmann2019}& 53.5& 52.3 & 19.5 & 36.6 & 12201 & 248047 & 2072\\
   LSST17\cite{FengMOT2019}& 54.7& 62.3 & 20.4 & 40.1 & 26091 & 228434 & \textbf{1243}\\
   FAMNet\cite{ChuFAM2019}& 52.0 & 48.7 & 19.1 & 33.4 & 14138 & 253616 &3072\\
   TT17\cite{ZhangTT2020}& 54.9 & \textbf{63.1} & 24.4 & 38.1 & 8895 & 233206 &2351\\
   Tractor++v2\cite{Bergmann2019}& 56.3 & 55.1 & 21.1 & 35.3 & \textbf{8666} & 235449 &1987\\
   CenterTrack\cite{zhou2020tracking}& 61.4& 53.3 & \textbf{27.9} & \textbf{31.4} & 15520 & 196886 &5326\\
   Unsup\cite{KarUnsup2020}& 61.7& 58.1 & 27.2 & 32.4 & 16872 & 197632 &1864\\
   ours& \textbf{62.0} & 59.5 & 27.8 & 31.5 & 15114 & \textbf{196672} &2621\\
   \bottomrule 
   \end{tabular}
   \caption{Result on MOT16 and MOT17 Public detection datasets.}
   \label{tab:MOTPub}
\end{table}

\textbf{Private Detection Result }
Our detector uses the detection result of Fair. Similarly, it can be seen from the table \ref{tab:MOTPri} that whether it is in MOT16 or MOT17 testing sets, among all the results, our MOTA, MT, ML and FN indexes are the best, and IDF1 and IDSw indexes are almost optimal. The MOTA metric achieves 76.5\%, exceeds Fair by 1.6 points on the MOT16; It achieves 76.2\%, exceeds Fair by 2.5 points and CenterTrack by 8.4 points on the MOT17.
\begin{table}
\scriptsize
	\centering
   \renewcommand\tabcolsep{3.5pt}
   \begin{tabular}{c|cccccccc}
   \toprule  
   \multicolumn{8}{c}{MOT16}\\
   \midrule
   Method& MOTA$\uparrow$ & IDF1$\uparrow$ & MT$\uparrow$  & ML$\downarrow$ & FP$\downarrow$ & FN$\downarrow$ & IDSw$\downarrow$\\
   \midrule  
   EAMTT\cite{SanEAM2016}& 52.5& 53.3 & 19.0 & 34.9 & \textbf{4407} & 81223 &910\\
   IOU\cite{BocIOU2017}& 57.1 & 46.9 & 23.6 & 32.9 & 5702 & 70278 &2167\\
   SORTwHPD16\cite{Bewleysort2016}& 59.8& 53.8 & 25.4 & 22.7 & 8698 & 63245 &1423\\
   DeepSORT\_2\cite{Wodeep2017}& 61.4& 62.2 & 32.8 & 18.2 & 12852 & 56668 & 781\\
   RAR16wVGG\cite{FangRAR2018}& 63.0& 63.8 & 39.9 & 22.1 & 13663 & 53248 & \textbf{482}\\
   POI\cite{YuPOI2016}& 66.1 & 65.1 & 34.0 & 20.8 & 5061 & 55914 &805\\
   Tube\_TK\_POI\cite{PangTube2020}& 66.9 & 62.2 & 39.0 & 16.1 & 11544 & 47502 &1236\\
   CTrackerV1\cite{PengCT2020}& 67.6 & 57.2 & 32.9 & 23.1 & 8934 & 48305 &1897\\
   Fair\cite{zhang2020simple}& 74.9 & \textbf{72.8} & 44.7 & 15.9 & 10163 &34484&1074\\
   ours& \textbf{76.5} & 68.6 & \textbf{52.8} & \textbf{12.3} & 12878 & \textbf{28982} &1026\\
   \toprule
   \multicolumn{8}{c}{MOT17}\\
   \midrule  
   Method& MOTA$\uparrow$ & IDF1$\uparrow$ & MT$\uparrow$  & ML$\downarrow$ & FP$\downarrow$ & FN$\downarrow$ & IDSw$\downarrow$\\
   \midrule  
   SCNet\cite{PangTube2020}& 60.0 & 54.4 & 34.4 & 16.2 & 72230 & 145851 &7611\\
   Tube\_TK\cite{PangTube2020}& 63.0 & 58.6 & 31.2 & 19.9 & 27060 & 177483 &4137\\
   CTrackerV1\cite{PengCT2020}& 66.6 & 57.4 & 32.2 & 24.2 & 22284 & 160491 &5529\\
   CTracker17\cite{zhou2020tracking}& 67.8 & 64.7 & 34.6 & 24.6 & \textbf{18498} & 160332 &\textbf{3039}\\
   Fair\cite{zhang2020simple}& 73.7 & \textbf{72.3} & 43.2 & 17.3 & 27507 &117477&3303\\
   ours& \textbf{76.2} & 68.0 & \textbf{51.1} & \textbf{13.6} & 32796 & \textbf{98475} &3237\\
   \bottomrule 
   \end{tabular}
   \caption{Result on MOT16 and MOT17 Private detection datasets.}
   \label{tab:MOTPri}
\end{table}

\subsection{Ablation Experiments}
\textbf{Feature influence} We first performed an ablation experiment on the selection of backbone without using any trick, i.e., associating match using similarity matrix composed of initial features and IOU. The results are shown in table \ref{tab:Ablation1}. Different neural networks have a certain impact on the accuracy of tracking. Simple network structures such as VGG16 and ResNet34 can have a very fast tracking speed, but the effect is not as good as deep networks such as ResNet101-FPN. The reason why ResNet101-fpn performs well is that it makes full use of the high-resolution information of low-level feature maps and high-level semantic information of high-level feature maps, this illustrates the important role of feature matching in multi-object tracking from the side. ResNet101-FPN is not used for detection, it simply extracts features, so it can achieve real-time tracking. Therefore, the backbone chosen for the Public and Private data sets in this article is Resnet101-fpn.
\begin{table}
\scriptsize
	\centering
   \begin{tabular}{c|ccccc}
   \toprule  
   Backbone& MOTA$\uparrow$ & IDF1$\uparrow$ & FP$\downarrow$ & FN$\downarrow$ & IDSw$\downarrow$\\
   \midrule  
   VGG16  & 56.6 & 55.1 & 4458 & 67446 & 1202 \\
   ResNet34  & 57.2 & 55.6 & 4371 & 66479 & 1245 \\
   ResNet50 & 58.8 & 57.1 & 4266 & 63966 & 1168\\
   ResNet101 & 59.7 & 58.5 & 4038 & 62672 &1174\\
   ResNet101-FPN & 61.2 & 59.1 & 3314 & 61057 &986\\
   \bottomrule 
   \end{tabular}
   \caption{Backbone experiments on public MOT17 train dataset. The ResNet101-FPN achieves the best result. 
   This illustrates the importance 
   of feature matching in multi-object tracking. 
   }
   \label{tab:Ablation1}
\end{table}

\textbf{Feature re-extraction} We add motion prediction, but tracklets do not re-extract feature in the current frame. We use four motion estimation methods, i.e., LK, pyramid LK, Kalman filter and FlowNet, the results are shown in the table \ref{tab:flow1}. In addition, we do a control experiment of re-extracting feature of tracklets in the current frame, as shown in the table \ref{tab:flow2}. We can see that whether it is the MOTA metric or the IDF1 metric, the pyramid LK method is slightly better than other methods, mainly because the pyramid LK method can estimate large emotion well. After tracklets re-extract feature in the current frame, the effect is greatly improved, which shows the importance of re-extracting feature on the same frame for the same object on different timestamps.
\begin{table}
\scriptsize
	\centering
   \begin{tabular}{c|ccccc}
   \toprule  
   Flow method& MOTA$\uparrow$ & IDF1$\uparrow$ & FP$\downarrow$ & FN$\downarrow$ & IDSw$\downarrow$\\
   \midrule  
   baseline  & 61.2 & 59.1 & 3314 & 61057 & \textbf{986} \\
    LK\cite{Bruce1981} & 62.1 & 60.1 & 3305 & 59378 & 1158 \\
   pyramid-LK\cite{Bouguet00pyramidalimplementation} & \textbf{63.2} & \textbf{63.1} & \textbf{3174} & \textbf{57682} & 1132\\
   Kalman filter\cite{Welch1995} & 62.4 & 62.2 & 3263 & 58994 &1079\\
   FlowNet\cite{Eddy2017flownet2} & 62.3 & 62.2 & 3436 & 58991 & 1077 \\
   \bottomrule 
   \end{tabular}
   \caption{Four types of optical flow method, tracklets do not re-extract features in the current frame, comprehensive performance of pyramid LK method is slightly better than other methods. We use the MOT17 training sets, half of the data is used as the training set, and the other half is used as the validation set. }
   \label{tab:flow1}
\end{table}

\begin{table}
\scriptsize
	\centering
   \begin{tabular}{c|ccccc}
   \toprule  
   Flow method& MOTA$\uparrow$ & IDF1$\uparrow$ & FP$\downarrow$ & FN$\downarrow$ & IDSw$\downarrow$\\
   \midrule  
    LK\cite{Bruce1981} & 63.1 & 61.8 & 3336 & 57665 & 1156 \\
   pyramid-LK\cite{Bouguet00pyramidalimplementation} & \textbf{65.4} & \textbf{65.3} & 3567 & \textbf{53671} & 1044\\
   Kalman filter\cite{Welch1995} & 63.9 & 63.7 & \textbf{3245} & 56547 &1017\\
   FlowNet\cite{Eddy2017flownet2} & 64.2 & 62.7 & 3489 & 55812 & \textbf{1003}\\
   \bottomrule 
   \end{tabular}
   \caption{Four types optical flow method, tracklets re-extract features in the current frame, it shows the importance of extracting feature on the same frame for the same object on different timestamps.}
   \label{tab:flow2}
\end{table}

\textbf{Information aggregation} We only add the AGNN part based on ResNet101-fpn, and use the MSE Loss. We compared three situations of whether to use AGNN and whether to use the adaptive part of AGNN, the result is shown in the table \ref{tab:GNN}. The tracking effect of using GNN but not adding the adaptive part is better than not using GNN, and the result of adding the adaptive part is better. This is because adaptive weights can fully learn the important parts of each object.
\begin{table}
\scriptsize
	\centering
   \begin{tabular}{c|ccccc}
   \toprule  
   Flow method& MOTA$\uparrow$ & IDF1$\uparrow$ & FP$\downarrow$ & FN$\downarrow$ & IDSw$\downarrow$\\
   \midrule  
   baseline  & 61.2 & 59.1 & 3314 & 61057 & \textbf{986} \\
    AGNN* & 62.3 & 61.3 & 3395 & 58962 & 1147 \\
   AGNN & \textbf{64.1} & \textbf{63.1} & \textbf{3268} & \textbf{55149} & 1055\\
   \bottomrule 
   \end{tabular}
   \caption{AGNN* means AGNN without adaptive part. We can get that GNN can fully use tracklets, location and appearance information to update feature vector, and the adaptive part can fully learn the important parts of each object. It's beneficial for re-identifying occluded objects.}
   \label{tab:GNN}
\end{table}

\textbf{Loss effect comparison} We compared the use of MSE Loss, Triplet Loss and BMSE Loss respectively  based on ResNet101-fpn, We only updates the parameters of the fully connected layer in training phase, the results are shown in the table \ref{tab:LOSS}. It can be seen that the results of BMSE is better than result of Triplet Loss, obviously simple BMSE Loss can replace Triplet Loss.
\begin{table}
\scriptsize
	\centering
   \begin{tabular}{c|ccccc}
   \toprule  
   Flow method& MOTA$\uparrow$ & IDF1$\uparrow$ & FP$\downarrow$ & FN$\downarrow$ & IDSw$\downarrow$\\
   \midrule  
   baseline  & 61.2 & 59.1 & 3314 & 61057 & \textbf{986} \\
    MSE & 61.5 & 60.8 & \textbf{3497} & 60144 & 1211 \\
   BMSE & \textbf{62.0} & \textbf{61.7} & 3768 & \textbf{59112} & 1129\\
   Triplet\cite{lu2020retinatrack} & 61.9 & \textbf{61.7} & 3743 & 59288 &1147\\
   \bottomrule 
   \end{tabular}
   \caption{We only updates the parameters of the fully connected layer in training phase, result shows that BMSE Loss performs better than Triplet Loss.}
   \label{tab:LOSS}
\end{table}

\textbf{Ablation study} We analyzed multiple combinations, the backbone used in table \ref{tab:Ablation2} is Resnet101-fpn. We do ablation experiments on whether to use feature alignment(it means tracklets re-extract features from the current frame), AGNN, and balanced MSE LOSS. It can be seen from table \ref{tab:Ablation2} that only using feature alignment, the MOTA index is 4.2 points higher than not using any trick, and 1.3 points higher than using AGNN only, and 2.4 points higher than using BMSE only. On the basis of using feature alignment, coupled with the AGNN updating feature, it is 2.7 points higher. Finally, with the addition of BMSE, the MOTA index is 0.8 points higher. It can be observed from table \ref{tab:Ablation2}, the feature alignment is a key point that makes the tracking performance the most improved. AGNN also has a very significant effect on improving tracking performance, and the BMSE performs well, too.

\begin{table}
\scriptsize
	\centering
	\renewcommand\tabcolsep{3.5pt}
   \begin{tabular}{|c|c|c|ccccc}
   \toprule  
\multicolumn{1}{c}{\begin{tabular}[c]{@{}c@{}}Feature\\ alignment\end{tabular}}& AGNN & BMSE & MOTA$\uparrow$ & IDF1$\uparrow$ & FP$\downarrow$ & FN$\downarrow$ & IDSw$\downarrow$\\
   \midrule  
    $\times$ & $\times$ & $\times$ & 61.2 & 59.1 & 3314 & 61057 &986\\
    $\surd$& $\times$ & $\times$ & 65.4 & 65.3 & 3567 & 53671 &1044\\
    $\times$& $\surd$ & $\times$ & 64.1 & 63.3 & 3268 & 55149 &1055\\
    $\times$& $\times$ & $\surd$ & 62.0 & 61.7 & 3768 & 59112 &1129\\
    $\surd$& $\surd$ & $\times$ & 68.1 & 65.8 & 3137 & 49664 &933\\
    $\surd$& $\times$ & $\surd$ & 66.3 & 65.8 & 3188 & 52565 &1013\\
    $\times$& $\surd$ & $\surd$ & 65.0 & 63.6 & 3347 & 54584 &1052\\
    $\surd$& $\surd$ & $\surd$ & \textbf{68.9} & \textbf{66.3} & \textbf{2897} & \textbf{48624} &\textbf{865}\\
   \bottomrule 
   \end{tabular}
   \caption{Ablation study.}
   \label{tab:Ablation2}
\end{table}

\textbf{Robustness of TPAGT} We use VGG16 and ResNet of other types as backbone for ablation experiments, and the results are shown in table \ref{tab:Ablation4}. We found that after using feature alignment, AGNN and balanced MSE LOSS, the influence of network structure on tracking accuracy is not so obvious. After each type of network uses feature alignment and AGNN, their MOTA metric is closer. Especially when VGG16 is used as the backbone, on the premise of using feature alignment, the effect of using AGNN is actually improved by 4.9 points, which is basically the same as the effect of other ResNet networks. On the one hand, it shows that AGNN is a good way to integrate global spatial-temporal location and appearance information , it means AGNN allows the same object on different frames to learn more similar feature vector. On the other hand, the AGNN part of TPAGT is robust. The features are extracted from different backbones, AGNN can fuse the location, appearance and historical information to updated the features, then these features of different objects can be very distinguishable, and the features of same object can be very similar.

\begin{table}
\scriptsize
	\centering
	\renewcommand\tabcolsep{3.5pt}
   \begin{tabular}{c|c|c|ccccc}
   \toprule  
   Backbone&  AGNN & BMSE & MOTA$\uparrow$ & IDF1$\uparrow$ & FP$\downarrow$ & FN$\downarrow$ & IDSw$\downarrow$\\
   \midrule  
   VGG16&   $\times$ & $\times$ & 62.8 & 61.7 & 3947 & 57610 & 1105\\
   ResNet34 & $\times$ & $\times$ & 63.0 & 62.0 & 3804 & 56170 & 1179 \\
   ResNet50 & $\times$ & $\times$ & 63.4 & 62.5 & 3845 & 56779 & 1027\\
   ResNet101 & $\times$ & $\times$ & 64.0 & 62.8 & 3761 & 56411 &1069\\
   VGG16 & $\surd$ & $\times$ & 67.7 & 65.3 & 3384 & 50011 &  1013\\
   ResNet34 & $\surd$ & $\times$ & 67.8 & 65.3 & 3199 & 49972 &  1067\\
   ResNet50 & $\surd$ & $\times$ & 67.8 & 65.3 & 3199 & 49972 & 1067\\
   ResNet101 & $\surd$ & $\times$ & 68.0 & 65.4 & 3147 & 49676 &1079\\
   VGG16 & $\surd$ & $\surd$ & 68.5 & 66.1 & 3112 & 48905 & 1043 \\
   ResNet34 & $\surd$ & $\surd$ & 68.6 & 66.1 & 3066 & 48914 & 912 \\
   ResNet50 & $\surd$ & $\surd$ & 68.6 & 66.1 & 3066 & 48914 & 912\\
   ResNet101 & $\surd$ & $\surd$ & 68.7 & 66.3 & 2996 & 48832 &895\\
   \midrule
   ResNet101-FPN & $\surd$ & $\surd$ & \textbf{68.9} & \textbf{66.3} & \textbf{2897} & \textbf{48624} &\textbf{865}\\
   \bottomrule 
   \end{tabular}
   \caption{Ablation experiment on the MOT17 training datasets, half of the data is used as the training set, and the other half is used as the validation set. Feature alignment method is used by default.}
   \label{tab:Ablation4}
\end{table}

\section{Conclusion}\label{section:Conclusion}

We know usual tracking methods extract the features of the tracklets from previous frames rather than the current frame. And actually a lot of information is not fully utilized in these methods. So we further study how to extract better features of the objects to obtain preferred matching.
In this work, we
propose the TPAGT tracker,
which is an end-to-end learning framework.
Our method predicts the motion of tracklets and extracts their features in the current frame, so that the features can be aligned. Then we input the features into Adaptive Graph Neural Network together with the detected ones. AGNN can integrate tracklets with location and appearance information to update the features. We also introduce the balanced MSE Loss during the training phase to obtain improved data distribution. FGAGT outperforms the state-of-the-art methods on Public and Private MOT Challenge datasets and shows great tracking accuracy.


\section*{Acknowledgement}
This work was supported by Major Scientifc Research Project of Zhejiang Lab (No. 2019DB0ZX01).


\begin{thebibliography}{99}

    \bibitem{zhou2020tracking}
    Zhou, Xingyi and Koltun, Vladlen and Kr{\"a}henb{\"u}hl, Philipp.:
    Tracking Objects as Points. In: ECCV(2020).

    \bibitem{zhang2020simple}
    Zhang, Yifu and Wang, Chunyu and Wang, Xinggang and Zeng, Wenjun and Liu, Wenyu.:
    A Simple Baseline for Multi-Object Tracking. In: arXiv preprint arXiv:2004.01888(2020).

    \bibitem{Bou2000Pyra}
    Bouguet, Jean-Yves Pyramidal.:
    Implementation of the Lucas Kanade Feature Tracker Description of the algorithm. In: Intel Corporation Microprocessor Research Labs (2000).

    \bibitem{Milan2016MOT}
    Milan, A., Leal-Taix$\acute{e}$, L., Reid, I., Roth, S. \& Schindler, K.:
    MOT16: A Benchmark for Multi-Object Tracking. In: arXiv preprint arXiv: 1603.00831(2016).

    \bibitem{Baker2011Date}
    Baker, S., Scharstein, D., Lewis, J.P. et al.:
    A Database and Evaluation Methodology for Optical Flow. In: Int J Comput Vis 92, 1-31(2011). https://doi.org/10.1007/s11263-010-0390-2

    \bibitem{Jianbo1994gft}
    Jianbo Shi, \& Tomasi.: Good features to track. In: Proceedings of IEEE Conference on Computer Vision and Pattern Recognition CVPR(1994). doi:10.1109/cvpr.1994.323794.

    \bibitem{Bouguet00pyramidalimplementation}
    Jean-yves Bouguet.:
    Pyramidal implementation of the Lucas Kanade feature tracker.
    In: Intel Corporation, Microprocessor Research Labs(2000).

    \bibitem{G. Farneback03}
    G. Farneb?ck.:
    Two-Frame Motion Estimation Based on Polynomial Expansion.In:
    Lecture Notes in Computer Science, pp. 363-370(2003).

    \bibitem{Tao2012}
    Tao, M., Bai, J., Kohli, P., Paris, S.:
    SimpleFlow: A Non-iterative, Sublinear Optical Flow Algorithm. In:
    Computer Graphics Forum, 31(2pt1), 345¨C353. (2012). doi:10.1111/j.1467-8659.2012.03013.x

    \bibitem{K.P.1981Determining}
    Berthold K.P. Horn, Brian G. Schunck.:
    Determining optical flow. In: Artificial Intelligence, Volume 17, Issues 1-3, Pages 185-203(1981), ISSN 0004-3702, https://doi.org/10.1016/0004-3702(81)90024-2

    \bibitem{Bruce1981}
    Bruce D. Lucas and Takeo Kanade.:
    An iterative image registration technique with an application to stereo vision. In: In Proceedings of the 7th international joint conference on Artificial intelligence - Volume 2 (IJCAI'81). Morgan Kaufmann Publishers Inc., San Francisco, CA, USA, 674-679(1981).

    \bibitem{Farneback2001ICCV}
    Farneback, G. (n.d.).:
    Very high accuracy velocity estimation using orientation tensors, parametric motion, and simultaneous segmentation of the motion field. In:
    Proceedings Eighth IEEE International Conference on Computer Vision. ICCV(2001). doi:10.1109/iccv.2001.937514

    \bibitem{Alexey2015flownet1}
    Alexey Dosovitskiy, Philipp Fischer, Eddy Ilg, P. H?usser, C. Haz?rba?, V. Golkov, P. Smagt, D. Cremers, Thomas Brox.: FlowNet: Learning Optical Flow with Convolutional Networks. In: IEEE International Conference on Computer Vision (ICCV), 2015.

    \bibitem{Eddy2017flownet2}
    Ilg Eddy, Mayer Nikolaus, Saikia Tonmoy, Keuper Margret, Dosovitskiy Alexey, Brox, Thomas.:
    FlowNet 2.0: Evolution of Optical Flow Estimation with Deep Networks. In: Conference: 2017 IEEE Conference on Computer Vision and Pattern Recognition (CVPR), 1647-1655(2017).

    \bibitem{Scarselli2009}
    Scarselli F , Gori M , Tsoi A C , et al. The Graph Neural Network Model[J]. IEEE Transactions on Neural Networks, 2009, 20(1):61.


    \bibitem{luo2014multiple}
    Wenhan Luo and Junliang Xing and Anton Milan and Xiaoqin Zhang and Wei Liu and Xiaowei Zhao and Tae-Kyun Kim.:
    Multiple Object Tracking: A Literature Review. In: arXiv, cs.CV(2014).

    \bibitem{Keni2008}
    Keni B , Rainer S . Evaluating Multiple Object Tracking Performance: The CLEAR MOT Metrics[J]. Eurasip Journal on Image \& Video Processing, 2008, 2008(1):246309.

    \bibitem{He2015}
    Ren S , He K , Girshick R , et al. Faster R-CNN: Towards Real-Time Object Detection with Region Proposal Networks[J]. IEEE Transactions on Pattern Analysis and Machine Intelligence, 2015, 39(6).

    \bibitem{H1995}
    H. W. Kuhn. The hungarian method for the assignment problem. Naval research logistics quarterly, 1955.

    \bibitem{Bewley2016}
    A. Bewley, Z. Ge, L. Ott, F. Ramos and B. Upcroft, "Simple online and realtime tracking," 2016 IEEE International Conference on Image Processing (ICIP), Phoenix, AZ, 2016, pp. 3464-3468, doi: 10.1109/ICIP.2016.7533003.

    \bibitem{Wojke2017}
    N. Wojke, A. Bewley and D. Paulus, "Simple online and realtime tracking with a deep association metric," 2017 IEEE International Conference on Image Processing (ICIP), Beijing, 2017, pp. 3645-3649, doi: 10.1109/ICIP.2017.8296962.

    \bibitem{Liu2018}
    K. Liu, Y. Shen and L. Chen, "Simple online and realtime tracking with spherical panoramic camera," 2018 IEEE International Conference on Consumer Electronics (ICCE), Las Vegas, NV, 2018, pp. 1-6, doi: 10.1109/ICCE.2018.8326132.

    \bibitem{Fu2019}
    Fu H., Wu L., Jian M., Yang Y., Wang X. (2019) MF-SORT: Simple Online and Realtime Tracking with Motion Features. In: Zhao Y., Barnes N., Chen B., Westermann R., Kong X., Lin C. (eds) Image and Graphics. ICIG 2019. Lecture Notes in Computer Science, vol 11901. Springer, Cham. https://doi.org/10.1007/978-3-030-34120-6\_13

    \bibitem{Sergey2019}
    Sergey Menshov, Yan Wang, Andrey Zhdanov, Eugene Varlamov, Dmitry Zhdanov, "Simple online and realtime tracking people with new ¡°soft-iou¡± metric," Proc. SPIE 11342, AOPC 2019: AI in Optics and Photonics, 113420M (18 December 2019); https://doi.org/10.1117/12.2547922

    \bibitem{Hou2019}
    X. Hou, Y. Wang and L. Chau, "Vehicle Tracking Using Deep SORT with Low Confidence Track Filtering," 2019 16th IEEE International Conference on Advanced Video and Signal Based Surveillance (AVSS), Taipei, Taiwan, 2019, pp. 1-6, doi: 10.1109/AVSS.2019.8909903.

    \bibitem{Maher2018}
    Maher, A., Taha, H. Zhang, B. Realtime multi-aircraft tracking in aerial scene with deep orientation network. J Real-Time Image Proc 15, 495¨C507 (2018). https://doi.org/10.1007/s11554-018-0780-1


    \bibitem{Caelles2016}
    S. Caelles Prat, ¡°Video Object Segmentation by Tracking Structured Key Points and Contours,¡± Projecte Final de M¨¤ster Oficial, UPC, Escola T¨¨cnica Superior d'Enginyeria de Telecomunicaci¨® de Barcelona, Departament de Teoria del Senyal i Comunicacions, 2016.

    \bibitem{Z2017}
    Yi Z , Tao X U , Dong X U , et al. Coordinating Multiple Cameras to Assist Tracking Moving Objects Based on Network Topological Structure[J]. Geomatics Information ence of Wuhan University, 2017, 42(8):1117-1122.

    \bibitem{Malinovskiy2011}
    Malinovskiy, Yegor Wu, Yao-Jan Wang, Y.. (2009). Video-Based Vehicle Detection and Tracking Using Spatiotemporal Maps. Transportation Research Record. 2121. 81-89. 10.3141/2121-09.

    \bibitem{Mittal2002}
    Mittal, A. . "M2Tracker: A Multi-View Approach to Segmenting and Tracking People in a Cluttered Scene." Proc.of European Conf.on Computer Vision (2002).

    \bibitem{Segen1996}
    Segen J . A camera-based system for tracking people in real time. In: International Conference on Pattern Recognition. IEEE, 1996.

    \bibitem{Khan2009}
    Khan S M , Shah M . Tracking Multiple Occluding People by Localizing on Multiple Scene Planes[J]. IEEE Transactions on Pattern Analysis and Machine Intelligence, 2009, 31(3):505-519.

    \bibitem{Cohen2018}
    Cohen, Isaac and Ayache, Nicholas and Sulger, Patrick.: Tracking points on deformable objects using curvature information. In: ECCV(1992), 458--466.


    \bibitem{zhang2018integrated}
    Zheng Zhang and Dazhi Cheng and Xizhou Zhu and Stephen Lin and Jifeng Dai.:
    Integrated Object Detection and Tracking with Tracklet-Conditioned Detection.
    In: arXiv:1811.11167 [cs.CV](2018).

    \bibitem{lu2020retinatrack}
    Zhichao Lu and Vivek Rathod and Ronny Votel and Jonathan Huang.:
    RetinaTrack: Online Single Stage Joint Detection and Tracking.
    In: arXiv:2003.13870 [cs.CV](2020).

    \bibitem{wang2020joint}
    Yongxin Wang and Xinshuo Weng and Kris Kitani.:
    Joint Detection and Multi-Object Tracking with Graph Neural Networks. In:
    arXiv:2006.13164 [cs.CV](2020).

    \bibitem{zhou2019objects}
    Xingyi Zhou and Dequan Wang and Philipp Kr?henb¨¹hl.:
    Objects as Points. In: arXiv:1904.07850 [cs.CV](2019).

    \bibitem{Linfocal2017}
    Lin T Y , Goyal P , Girshick R , et al. Focal Loss for Dense Object Detection[J]. IEEE Transactions on Pattern Analysis \& Machine Intelligence, 2017, PP(99):2999-3007.

    \bibitem{Welch1995}
    Welch, G., Bishop, G., et al.: An introduction to the kalman filter (1995)









    \bibitem{Ho1997}
    S. Hochreiter and J. Schmidhuber. 1997. Long short-term memory. Neural Computation,
    9(8):1735¨C1780. DOI: 10.1162/neco.1997.9.8.1735.
















    \bibitem{Bernardinclear2008}
    Bernardin, K. \& Stiefelhagen, R. Evaluating Multiple Object Tracking Performance: The CLEAR MOT Metrics. Image and Video Processing, 2008(1):1-10, 2008

    \bibitem{Ristani2016}
    Ristani, E., Solera, F., Zou, R., Cucchiara, R. \& Tomasi, C. Performance Measures and a Data Set for multi-object, Multi-Camera Tracking. In ECCV workshop on Benchmarking multi-object Tracking, 2016.

    \bibitem{Philipp2019}
    Philipp Bergmann, Tim Meinhardt, and Laura Leal-Taixe.
    Tracking without bells and whistles. arXiv preprint
    arXiv:1903.05625, 2019.

    \bibitem{FangPAR2018}
    Fang, K., Xiang, Y., Li, X., Savarese, S.: Recurrent autoregressive networks for
    online multi-object tracking. In: W ACV. pp. 466¨C475 (2018).

    \bibitem{SadeAMIR2017}
    Sadeghian, A., Alahi, A., Savarese, S.: Tracking the untrackable: Learning to track
    multiple cues with long-term dependencies. In: ICCV. pp. 300¨C311 (2017).

    \bibitem{ChuKCF2019}
     Chu, P., Fan, H., Tan, C.C., Ling, H.: Online multi-object tracking with instance-
    aware tracker and dynamic model refreshment. In: W ACV. pp. 161¨C170 (2019)

    \bibitem{ChenMOT2018}
     Chen, L., Ai, H., Zhuang, Z., Shang, C.: Real-time multiple people tracking with
    deeply learned candidate selection and person re-identification. In: ICME. pp. 1¨C6
    (2018).

    \bibitem{XuSTRN2019}
     Xu, J., Cao, Y., Zhang, Z., Hu, H.: Spatial-temporal relation networks for multi-
     object tracking. In: ICCV. pp. 3988¨C3998 (2019).

    \bibitem{Xudeep2020}
    Xu, Y., Osep, A., Ban, Y., Horaud, R., Leal-Taix e, L., Alameda-Pineda, X.: How
    to train your deep multi-object tracker. In: CVPR (2020).

    \bibitem{GuMPN2020}
    G. Bras¨® and L. Leal-Taix¨¦, "Learning a Neural Solver for Multiple Object Tracking," 2020 IEEE/CVF Conference on Computer Vision and Pattern Recognition (CVPR), Seattle, WA, USA, 2020, pp. 6246-6256, doi: 10.1109/CVPR42600.2020.00628.

    \bibitem{Hornakova2020}
    Hornakova, A., Henschel, R., Rosenhahn, B., and Swoboda, P.: 2020, arXiv e-prints, arXiv:2006.14550.

    \bibitem{KarUnsup2020}
    Karthik S , Prabhu A , Gandhi V . Simple Unsupervised Multi-Object Tracking[J]. 2020.

    \bibitem{Bergmann2019}
    Bergmann, P ., Meinhardt, T., Leal-Taixe, L.: Tracking without bells and whistles. In: ICCV(2019).

    \bibitem{FengMOT2019}
     Feng, W., Hu, Z., Wu, W., Yan, J., Ouyang, W.: Multi-object tracking with multiple cues and
     switcher-aware classification. arXiv:1901.06129 (2019).

    \bibitem{ZhuDMAN2018}
    Zhu, J., Yang, H., Liu, N., Kim, M., Zhang, W., Yang, M.H.: Online multi-object
    tracking with dual matching attention networks. In: ECCV. pp. 366¨C382 (2018).


    \bibitem{ChenMOTDT2018}
     Chen, L., Ai, H., Zhuang, Z., Shang, C.: Real-time multiple people tracking with
    deeply learned candidate selection and person re-identification. In: ICME. pp. 1¨C6(2018).

    \bibitem{ChuFAM2019}
    Chu, P., Ling, H.: Famnet: Joint learning of feature, affinity and multi-dimensional
    assignment for online multiple object tracking. In: ICCV. pp. 6172¨C6181 (2019).

    \bibitem{ZhangTT2020}
    Y. Zhang, H. Sheng, Y. Wu, S. Wang, W. Lyu, W. Ke, Z. Xiong. Long-term Tracking with Deep Tracklet Association. In IEEE Transactions on Image Processing, 2020.

    \bibitem{SanEAM2016}
    Sanchez-Matilla, R., Poiesi, F., Cavallaro, A.: Online multi-object tracking with
    strong and weak detections. In: European Conference on Computer Vision. pp.
    84¨C99. Springer (2016).

    \bibitem{BocIOU2017}
    E. Bochinski, V. Eiselein, T. Sikora. High-Speed Tracking-by-Detection Without Using Image Information. In International Workshop on Traffic and Street Surveillance for Safety and Security at IEEE AVSS 2017, 2017.

    \bibitem{Bewleysort2016}
    Bewley, A., Ge, Z., Ott, L., Ramos, F., Upcroft, B.: Simple online and realtime
    tracking. In: 2016 IEEE International Conference on Image Processing (ICIP). pp.
    3464¨C3468. IEEE (2016).

    \bibitem{Wodeep2017}
    Wojke, N., Bewley, A., Paulus, D.: Simple online and realtime tracking with a deep
    association metric. In: 2017 IEEE international conference on image processing
    (ICIP). pp. 3645¨C3649. IEEE (2017).

    \bibitem{FangRAR2018}
    Fang, K., Xiang, Y., Li, X., Savarese, S.: Recurrent autoregressive networks for
    online multi-object tracking. In: 2018 IEEE Winter Conference on Applications of
    Computer Vision (W ACV). pp. 466¨C475. IEEE (2018).

    \bibitem{YuPOI2016}
    Yu, F., Li, W., Li, Q., Liu, Y., Shi, X., Yan, J.: Poi: Multiple object tracking with
    high performance detection and appearance feature. In: European Conference on
    Computer Vision. pp. 36¨C42. Springer (2016).


    \bibitem{PangTube2020}
    B. Pang, Y. Li, Y. Zhang, M. Li, C. Lu. TubeTK: Adopting Tubes to Track Multi-Object in a One-Step Training Model. In CVPR, 2020.

    \bibitem{PengCT2020}
    J. Peng, C. Wang, et.al. Chained-Tracker: Chaining Paired Attentive Regression Results for End-to-End Joint Multiple-Object Detection and Tracking. In ECCV Spotlight, 2020.

    \bibitem{He2018ROIAlign}
    Kaiming He and Georgia Gkioxari and Piotr Doll¨¢r and Ross Girshick.
    Mask R-CNN.
    In arXiv:1703.06870v3, cs.CV, 2018.

\end{thebibliography}
\end{document}